\documentclass[sigconf]{acmart}
\usepackage[utf8]{inputenc} %
\usepackage[T1]{fontenc}    %
\usepackage{url}            %
\usepackage{booktabs}       %
\usepackage{amsfonts}       %
\usepackage{nicefrac}       %
\usepackage{microtype}      %
\usepackage{lipsum}
\usepackage{tikz}
\usetikzlibrary{shapes, arrows, patterns} %
\usepackage{graphicx}       %
\graphicspath{{media/}}     %
\usepackage{multirow}
\usepackage{subcaption}
\usepackage{tabularx,xcolor}
\tikzstyle{box} = [rectangle, minimum width=3cm, minimum height=1cm, text centered, draw=black]
\tikzstyle{block} = [rectangle, draw, node distance=5cm, text width=3.5cm, text centered, rounded corners, minimum height=2em]
\tikzstyle{line} = [draw, -latex']
\definecolor{light-gray}{gray}{0.7}
\AtBeginDocument{%
  }

\copyrightyear{2025}
\acmYear{2025}
\setcopyright{cc}
\setcctype{by}
\acmConference[KDD '25]{Proceedings of the 31st ACM SIGKDD Conference on Knowledge Discovery and Data Mining V.1}{August 3--7, 2025}{Toronto, ON, Canada}
\acmBooktitle{Proceedings of the 31st ACM SIGKDD Conference on Knowledge Discovery and Data Mining V.1 (KDD '25), August 3--7, 2025, Toronto, ON, Canada}
\acmDOI{10.1145/3690624.3709436}
\acmISBN{979-8-4007-1245-6/25/08}

\begin{document}
\pagestyle{plain}

\title{Explainable LiDAR 3D Point Cloud Segmentation and Clustering for Detecting Airplane-Generated Wind Turbulence}

\author{Zhan Qu}
\affiliation{%
  \institution{Karlsruhe Institute of Technology}
  \city{Karlsruhe}
  \country{Germany}
}
\affiliation{%
  \institution{ScaDS.AI, TU Dresden}
  \city{Dresden}
  \country{Germany}
}
\email{zhan.qu@kit.edu}

\author{Shuzhou Yuan}
\affiliation{%
  \institution{ScaDS.AI, TU Dresden}
  \city{Dresden}
  \country{Germany}
}
\email{shuzhou.yuan@tu-dresden.de}

\author{Michael F{\"a}rber}
\affiliation{%
  \institution{ScaDS.AI, TU Dresden}
  \city{Dresden}
  \country{Germany}}
\email{michael.faerber@tu-dresden.de}

\author{Marius Brennfleck}
\affiliation{%
  \institution{Karlsruhe Institute of Technology}
  \city{Karlsruhe}
  \country{Germany}
}
\email{marius.brennfleck@student.kit.edu}

\author{Niklas Wartha}
\affiliation{%
 \institution{Institute of Atmospheric Physics, German Aerospace Center}
 \city{Oberpfaffenhofen}
 \country{Germany}}
\affiliation{%
 \institution{RWTH Aachen University}
 \city{Aachen}
 \country{Germany}}
\email{niklas.wartha@dlr.de}

\author{Anton Stephan}
\affiliation{%
  \institution{Institute of Atmospheric Physics, German Aerospace Center}
  \city{Oberpfaffenhofen}
  \country{Germany}}
\email{anton.stephan@dlr.de}

\renewcommand{\shortauthors}{Zhan Qu et al.}

\begin{abstract}
Wake vortices—strong, coherent air turbulences created by aircrafts—pose a significant risk to aviation safety and therefore require accurate and reliable detection methods. In this paper, we present an advanced, explainable machine learning method that utilizes Light Detection and Ranging (LiDAR) data for effective wake vortex detection. Our method leverages a dynamic graph CNN (DGCNN) with semantic segmentation to partition a 3D LiDAR point cloud into meaningful segments. Further refinement is achieved through clustering techniques. A novel feature of our research is the use of a perturbation-based explanation technique, which clarifies the model’s decision-making processes for air traffic regulators and controllers, increasing transparency and building trust. Our experimental results, based on measured and simulated LiDAR scans compared against four baseline methods, underscore the effectiveness and reliability of our approach. This combination of semantic segmentation and clustering for real-time wake vortex tracking significantly advances aviation safety measures, ensuring that these are both effective and comprehensible. 
\end{abstract}



\maketitle

\section{Introduction}

\begin{figure}[tb]
    \centering
    \begin{subfigure}{0.36\textwidth} %
        \includegraphics[width=\textwidth]{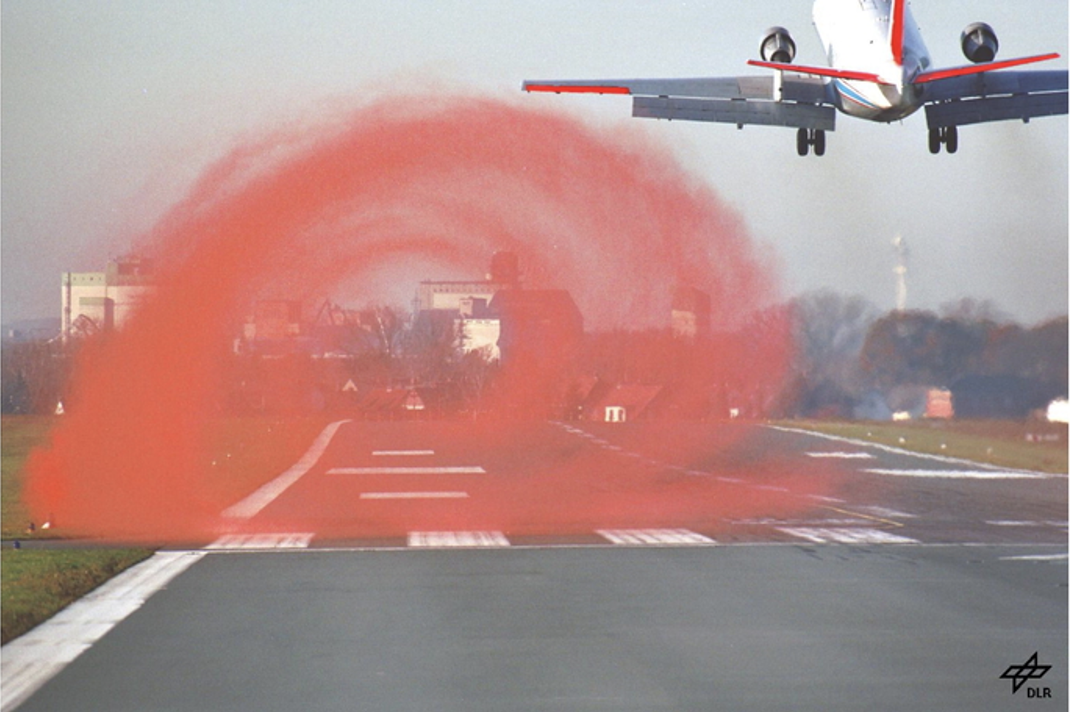}
    \end{subfigure}
    \hfill
    \begin{subfigure}{0.36\textwidth}
        \includegraphics[width=\textwidth]{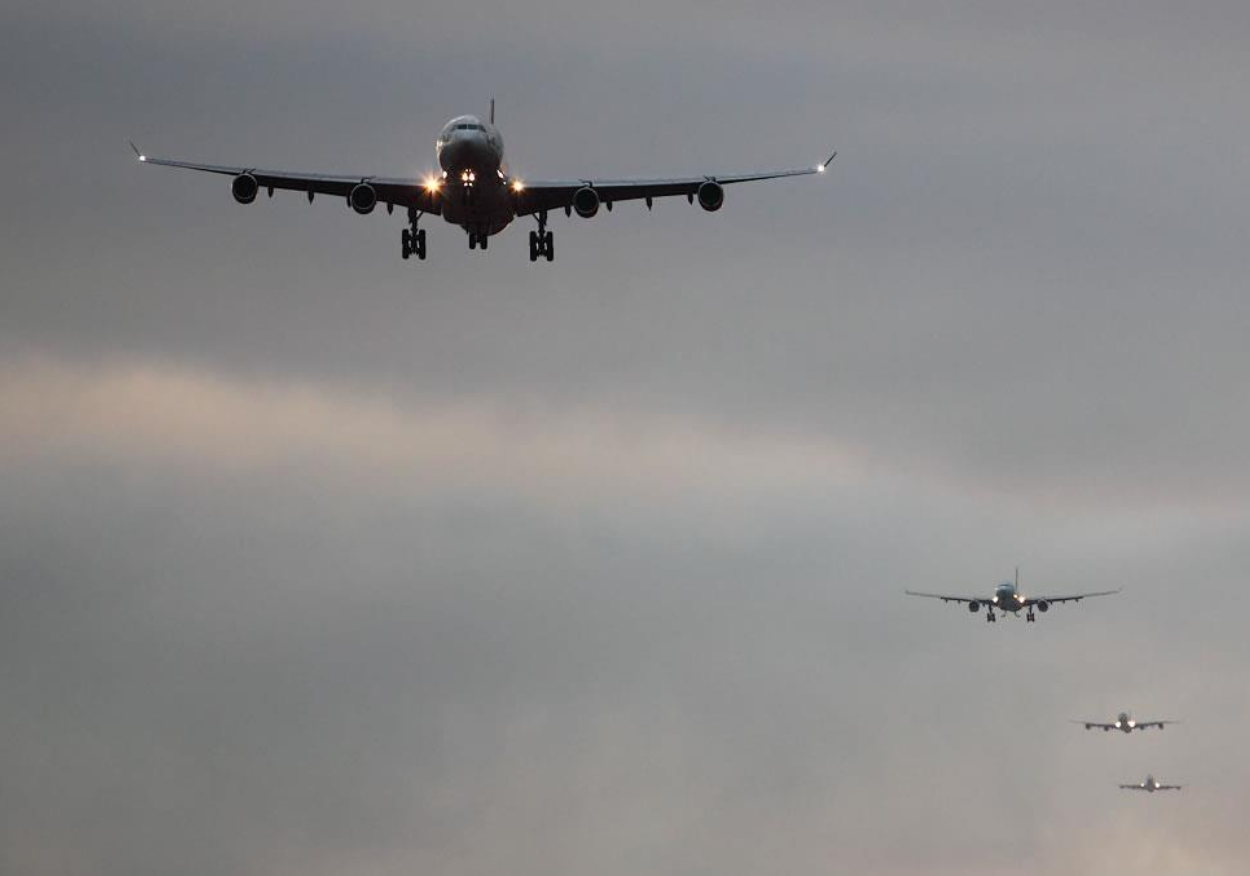}
    \end{subfigure}
    \caption{Top: Colored wake vortex generated by an aircraft. 
    Bottom: Aircrafts maintaining a minimum distance during landing to avoid wake vortices.} 
    \label{fig:wake vortices}
\end{figure}

As an inherent by-product of aerodynamic lift, every aircraft creates a pair of counter-rotating turbulent airflows, known as wake vortices (see Fig.~\ref{fig:wake vortices}), which pose a danger to subsequent air traffic, especially during the landing phase \cite{hallock2018review}. These vortices can persist for several minutes, creating hazardous conditions for aircraft that follow too closely. To mitigate this risk, aviation authorities have established minimum separation distances between landing aircrafts. However, this necessary precaution limits the operational capacity of airports, particularly those serving major metropolitan areas where air traffic demand is high. The rules regarding minimum separation distances are effective in theory, but they lack a data-driven foundation. In reality, some may be too strict or too loose, which may influence the efficiency of the airports \cite{holzapfel2021mitigating}. 

The challenge of wake vortices is exacerbated by the global growth in air traffic. Forecasts for European airports indicate a significant capacity deficit \cite{eurocontrol2018european}. By 2040, around 1.5 million flights will not be able to be handled, which corresponds to 8\% of total demand. This scenario could prevent 160 million passengers from traveling, resulting in a potential economic loss of around 88 billion euros for the European economy \cite{eurocontrol2018european,holzapfel2021mitigating}. Such figures underline the urgency of developing new approaches to air traffic management that can safely increase airport throughput without jeopardizing passenger safety.

In response to these challenges, researchers have turned to machine learning (ML) technologies to improve wake vortex detection and management. Recent studies \cite{weijun2019deep,wartha2022characterizing,pan2022identification} have demonstrated the feasibility of using LiDAR (Light Detection and Ranging) scans to detect wake vortices, employing ML algorithms to provide initial location estimates via a bounding box approach. These preliminary findings are encouraging, suggesting that ML, and Artificial Neural Networks (ANNs) in particular, could be applied more broadly within the field. The potential applications include not only the detection and qualitative analysis of wake vortices but also the precise quantification of their characteristics, such as their exact location and circulation (rotation of the air) \cite{weijun2019deep,weijun2019research,baranov2021wake,wartha2022characterizing}. Particularly in comparison to traditional physical wake vortex characterization algorithms for LiDAR measurements, the ML approaches offer far shorter processing times, enabling real-time applications \cite{wartha2022characterizing}.

Although ML and ANNs show promise for wake vortex detection, there are concerns about their effectiveness, efficiency, and transparency. Previous approaches often convert LiDAR scans into images \cite{weijun2019deep,wartha2022characterizing,stephan2023artificial}, which compromises the ability for efficient real-time detection. This method does not take advantage of the three-dimensional nature of LiDAR point clouds, leading to inefficiencies. Therefore, models specifically designed for 3D point clouds are preferred and may offer higher accuracy and efficiency than those adapted for image data. Furthermore, deep neural networks typically work as a ``black box,'' i.e. the reasons for their decisions are not recognizable to the human user. This opacity poses a significant challenge, particularly in safety-critical applications \cite{wartha2022characterizing,roadmap2021easa}, as users cannot directly understand the decision-making mechanisms of the models. Traditional approaches, which rely on converted images from point clouds, offer limited scope for interpretation and generalizability, since the structural information of the point clouds is omitted. It is not clear which parts or characteristics of the vortices are significant for the domain expert. A 3D point cloud approach may lead to a better understanding of the underlying flow field and may be generalized to other kinds of turbulence, like the background turbulence of the wind field. The need for ML systems to be effective, efficient, generalizable and interpretable cannot be overstated, as this is crucial for their acceptance and use in environments where safety is a top priority.

This paper, with real measurement data and synthetic data, aims to enhance the performance and interpretability of ML models for wake vortex detection. With our novel approach, we improve the effectiveness, efficiency, and explainability of wake vortex detection to substantially improve the safety and efficiency of air traffic control. Our approach, based on Dynamic Graph CNNs (DGCNNs) \cite{DGCNN_WSLSBS19}, is the first application of 3D point cloud segmentation specifically for wake vortex detection, setting a new standard in this field. We also present the first perturbation-based explanation method for 3D point cloud segmentation in wake vortex detection. This dual advance not only improves real-time decision making capabilities for air traffic control, but also offers the potential for widespread application with the possibility of being deployed at all airports and for flight connections worldwide. 
In extensive experiments using both unique real measurement data from the Vienna International Airport and synthetic data, we demonstrate that our approach surpasses all four baselines, including CNN \cite{lecun1998gradient}, YOLO \cite{redmon2016you}, PointNet \cite{PointNet_QSMG17}, and PointNet++ \cite{PointNet++_QYSG17}. 
Overall, by enabling more accurate detection and deeper understanding of wake vortices, we contribute to safer and more efficient flight operations. %

All in all, in this paper, we make the following contributions\footnote{The source code are available at: \href{https://anonymous.4open.science/r/WakeVortexDetection-D2D8/}{https://anonymous.4open.science/r/WakeVortexDetection-D2D8/}}:
\begin{itemize}
    \item We introduce the first application of 3D Point Cloud segmentation models (e.g., DGCNN) for wake vortex detection, establishing a new benchmark in the field. These models operate directly on 3D point clouds without requiring additional preprocessing steps, such as conversion to images. Our results demonstrate the superiority of these models compared to traditional image processing models (e.g., CNN). Recall is increased by up to 5.5\%, and the mean error is reduced by up to 12.35 meters.
    \item We integrate clustering algorithms (e.g., Agglomerative) with the segmentation models to further refine the results. 
    \item We develop the first perturbation-based explanation method for 3D point cloud segmentation in wake vortex detection. 
    \item We rigorously evaluate our methodology on both real-world data and synthetic data. 
    Our findings demonstrate that our approach enhances the accuracy, efficiency, and reliability of wake vortex detection.
\end{itemize}

\section{Background}

\subsection{LiDAR in Atmospheric Remote Sensing}

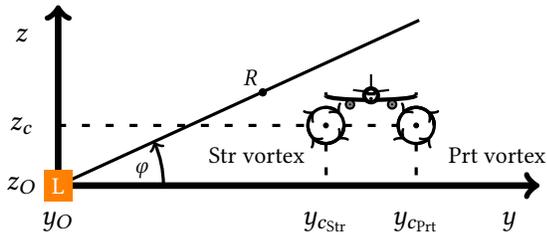
\begin{figure}[tb] %
    \centering
    \begin{tikzpicture}[scale=0.8]
    
        \draw[very thick, black, ->] (1.75,0) arc[start angle=0, end angle=24.624, radius=1.75];
        \node[] at (1.4,0.25)   (u) {$\varphi$};
        
        \draw[very thick, black] (0,0) -- (6,2.75);
        \draw[black,fill=black] (3.4,1.55) circle (0.06);
        \node[] at (3.2,1.8)   (s) {$R$};

        \draw[line width=2.5,->] (0,0) -- (8,0); 
        \node[] at (7.5, -0.6)   (a) {\Large $y$};
        \draw[line width=2.5,->] (0,0) -- (0,3); 
        \node[] at (-0.6, 2.5)   (b) {\Large $z$};
        
        \node[] at (-0.6, 0)   (c) {\Large $z_{O}$};
        \node[] at (0, -0.6)   (d) {\Large $y_{O}$};

        \node[white, fill=orange] at (0,0)  (h) {L};
        
        \draw[very thick, ->] (6.25,1) arc (0:140:0.3);
        \draw[very thick, ->] (6.25,1) arc (0:50:0.3);
        \draw[very thick, ->] (6.25,1) arc (0:230:0.3);
        \draw[very thick, ->] (6.25,1) arc (0:320:0.3);
        \draw[very thick] (6.25,1) arc (0:-95:0.3);
        
        \draw[black,fill=black] (5.95,1) circle (0.04);
        
        \node[] at (7.30,0.5)  (z) {Prt vortex};
        
        \draw[black,loosely dashed,very thick] (0,1) -- (4.15,1);
        \draw[black,loosely dashed,very thick] (4.75,1) -- (5.65,1);
        \node[] at (-0.6,1)  (w) {\Large $z_{c}$};
        
        \draw[black,loosely dashed,very thick] (5.95,0.7) -- (5.95,0);
        \node[] at (5.95,-0.6)  (v) {\Large $y_{c_{\textrm{Prt}}}$};

        \draw[very thick, ->] (4.75,1) arc (0:-320:0.3);
        \draw[very thick, ->] (4.75,1) arc (0:-230:0.3);
        \draw[very thick, ->] (4.75,1) arc (0:-140:0.3);
        \draw[very thick, ->] (4.75,1) arc (0:-50:0.3);
        \draw[very thick] (4.75,1) arc (0:95:0.3);
        
        \draw[black,fill=black] (4.45,1) circle (0.04);
        
        \node[] at (3.3,0.5)  (y) {Str vortex};
        
        \draw[black,loosely dashed,very thick] (4.45,0.7) -- (4.455,0);
        \node[] at (4.45,-0.6)  (r) {\Large $y_{c_{\textrm{Str}}}$};
        
            \draw[ultra thick] (4.45,1.5) .. controls (4.45,1.4) and (5.9,1.4) .. (5.95,1.5);
            \draw[very thick] (4.45,1.5) -- (4.45,1.55);
            \draw[very thick] (5.95,1.5) -- (5.95,1.55);
            \draw[thick] (5.2,1.5) -- (5.2,1.85);
            \draw[thick] (4.9,1.59) .. controls (5.15,1.555) and (5.25,1.555) .. (5.5,1.59);
            \draw[ultra thick,black,fill=white] (5.2,1.5) circle (0.115);
            \draw[very thick,black] (5.1186,1.5813) -- (5.15,1.525) -- (5.25,1.525) -- (5.2813,1.5813);
            \draw[black,fill=black] (5.2,1.445) circle (0.02);
            \draw[thick,black,fill=gray] (4.85,1.35) circle (0.07);
            \draw[black,fill=black] (4.85,1.35) circle (0.007);
            \draw[thick,black,fill=gray] (5.55,1.35) circle (0.07);
            \draw[black,fill=black] (5.55,1.35) circle (0.007);

    \end{tikzpicture}
    \caption{Schematic of LiDAR (L) scan measuring wake vortices of a landing aircraft (flying out of the page) perpendicular to the runway. Starboard (Str) and port (Prt) vortices are visible \cite{wartha2022characterizing}.}
    \label{fig:wake-vortices-slices}
    \Description[An illustration of wake vortices]{Schematic of LiDAR (L) scan measuring wake vortices of a landing aircraft}
\end{figure}

Light Detection and Ranging (LiDAR) instruments are versatile tools for atmospheric remote sensing. Capturing wind velocities, turbulence, and also wake vortices has become increasingly popular. Wake vortex detection in clear air conditions, especially, is typically performed via LiDAR, primarily due to its scattering advantages with aerosols (air particles) over traditional technologies such as radar, which feature longer wavelengths and are thus commonly employed for rainy conditions \cite{jianbing2017survey}. As a consequence of their long wavelength, radar can be limited in resolution. In contrast, LiDAR instruments feature a far shorter wavelength and thus increase the spatial resolution of precision measurements, which is ideal for identifying coherent structures such as wake vortices. Doppler wind LiDARs measure the radial component of the underlying wind field, including the fine-scale structures of wake vortices. LiDAR operates by emitting laser pulses and measuring the phase shift of the back-scattered signal from aerosols in their path. The Doppler shift yields the radial velocity of the fluid. This mechanism allows for the creation of two-dimensional radial velocity slices or scans of the atmosphere (see schematic in Fig.~\ref{fig:wake-vortices-slices}). The capacity of LiDAR to capture high-resolution spatial data is particularly useful for aviation safety, addressing the challenge of detecting wake vortices. The technology's ability to precisely and in real-time determine the strength and location of wake vortices makes it an essential tool for Wake Vortex Advisory Systems, thereby significantly maintaining the safety and enhancing the efficiency of aviation operations.

\subsection{Wake Vortex Formation and Ground Effects}

\begin{figure}[tb]
 \centering 
    \includegraphics[width=7.5cm]{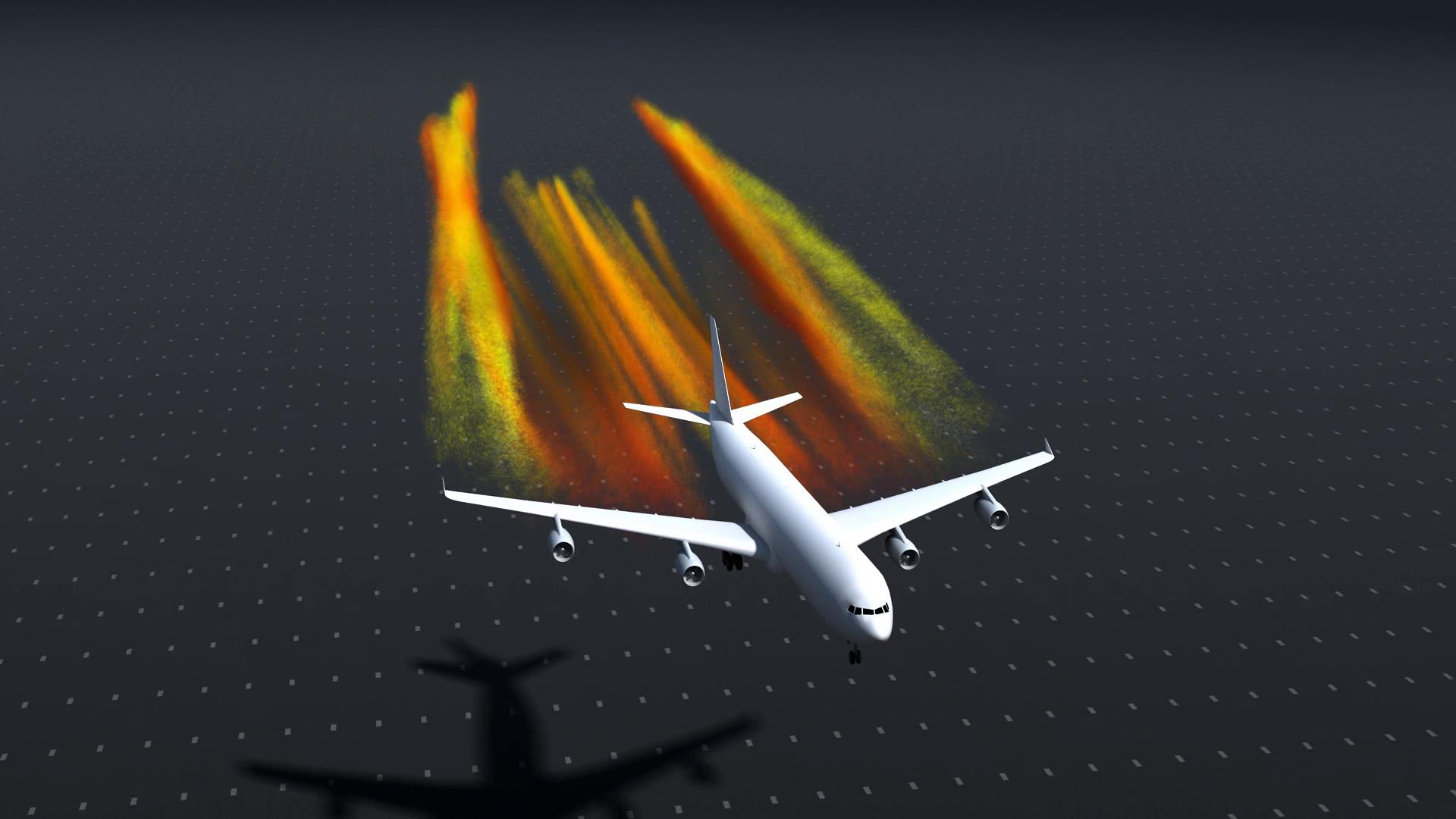}
    \caption{Large eddy simulation of the roll-up process for an A340 aircraft. Multiple vortices shed from the aircraft wings form a pair of two counter-rotating swirls.}
    \label{fig:wake-vortices-theory} %
    \Description[A simulation of wake vortices generated by an aircraft]{Large eddy simulation of the roll-up process for an A340 aircraft. Multiple vortices shed from the aircraft wings form a pair of two counter-rotating swirls}
\end{figure}

As a direct consequence of its lift generation, aircrafts produce wake vortices. Beneath the wings of an aircraft pressure is lower than above, leading to fluid acceleration around the wing and consequently vortex sheets shedding off and rolling up \cite{KUNDU2002629}. Primarily, vortices shed off the wing and flap tips of the aircraft \cite{stephan2019effects} as visualized in Fig.~\ref{fig:wake-vortices-theory}. The wake behind an aircraft, particularly when observed over long periods of time, develops into a highly complex and multiscale problem. The problem increases further in complexity 
near the ground, as is the case when airplanes land. This leads to the vortices descending and divering in a hyperbolic fashion \cite{robins2001algorithm}. Furthermore, they produce a boundary layer on the ground \cite{harvey1971flowfield}, ultimately leading to the detachment of secondary vortex structures which interact with the primary aircraft vortices. This interaction results in the rebound of the wake vortices, as well as potential stall over the runway when a weak crosswind is present \cite{spalart2001modeling,zheng1996study,holzapfel2007aircraft}. In this paper, LiDAR measurements include various atmospheric conditions and wake vortex stages.

\section{Related Work}

\subsection{Machine Learning for Wake Vortex Detection}
\label{subsection:ml_wv}
Several machine learning techniques have been considered to detect wake vortices in LiDAR scans. Specifically, traditional machine learning methods such as Support Vector Machine (SVM) and random forest have been applied to recognize the patterns in the LiDAR scans \cite{pan2020identification,weijun2021aircraft,pan2022recognition}. They captured the superficial characteristics of the LiDAR scans and made predictions based on these characteristics. With the development of deep neural networks, researchers have converted the LiDAR scans to images and leveraged Convolutional Neural Network (CNN), Long Short-Term Memory (LSTM), or the combination of CNN and LSTM to segment the wake vortices on the images and locate the positions of wake vortices \cite{abras2021application,ai2021deep,chu2024assessment,wartha2022characterizing}, leading to superior results compared to traditional machine learning methods.

More recently, advanced computer vision models such as YOLO and VGG have also been utilized to segment the wake vortices after the LiDAR scans are converted to images \cite{weijun2019deep,pan2022identification,stietz2022artificial,stephan2023artificial}. Despite the improvements brought by these models, the effort required to convert LiDAR scans to images entails additional pre-processing. Moreover, converting LiDAR scans to images may lead to the loss of vital information present in the original LiDAR scans, such as 
the interactions between the measured points, 
thus limiting the explainability of previous studies. 
Conversely, our method utilizes 3D point cloud data, enabling the model to directly read and process the LiDAR scans, thereby capturing the latent features represented within the LiDAR scan itself and making the prediction process more efficient.

\subsection{3D Point Cloud Analysis}

To our knowledge, 3D point cloud analysis techniques have yet to be applied to wake vortex detection, despite the inherently three-dimensional nature of the data (two dimensions in space, one radial velocity component). A point cloud is a collection of data points plotted in three-dimensional space, representing the physical contours of an environment or object. This representation is facilitated by laser scanning technologies, such as LiDAR, where each point in the cloud corresponds to a precise spatial location and can carry additional attributes like color, intensity, or in our case radial velocity. 
Point clouds are extensively applied across various domains, including autonomous driving, drones, and AR/VR techniques, leveraging their detailed spatial information for advanced functionality and analysis \cite{AutoDrive_CMWLX17}.

In the realm of 3D point cloud processing, several key tasks emerge, such as classification, segmentation, and object detection. Classification involves assigning semantic labels to shapes or objects depicted in the point clouds, such as identifying an entity as a vehicle, animal, or human \cite{Survey3D_GWHLLB21}. Segmentation tasks include part segmentation, dividing an object or scene into its constituent elements, and semantic segmentation, labeling every point with a specific type of object or surface for in-depth scene analysis \cite{Survey3D_GWHLLB21}. Object detection focuses on pinpointing and framing particular objects within the scene \cite{Survey3D_GWHLLB21}. 

The field has seen the development of various deep learning approaches tailored for the above tasks of point cloud analysis. Notably, PointNet \cite{PointNet_QSMG17} and its successor, PointNet++ \cite{PointNet++_QYSG17}, have been used by processing points individually or in hierarchical groupings for classification and segmentation tasks. Similarly, graph-based methods like Dynamic Graph Convolutional Neural Network (DGCNN) \cite{DGCNN_WSLSBS19} leverage the natural graph structure of point clouds for feature extraction and propagation, demonstrating effectiveness in segmentation and classification.

The application of deep learning to point clouds aims to address the challenges posed by their irregular structure, variable density, and inherent noise. 
The application of 3D point cloud segmentation techniques to wake vortex detection represents a novel frontier in the field. Despite the extensive use of these methods in various domains, their potential for identifying and analyzing wake vortices has yet to be explored.

\begin{figure*}[tb]
  \centering
   \includegraphics[width=0.95\textwidth]{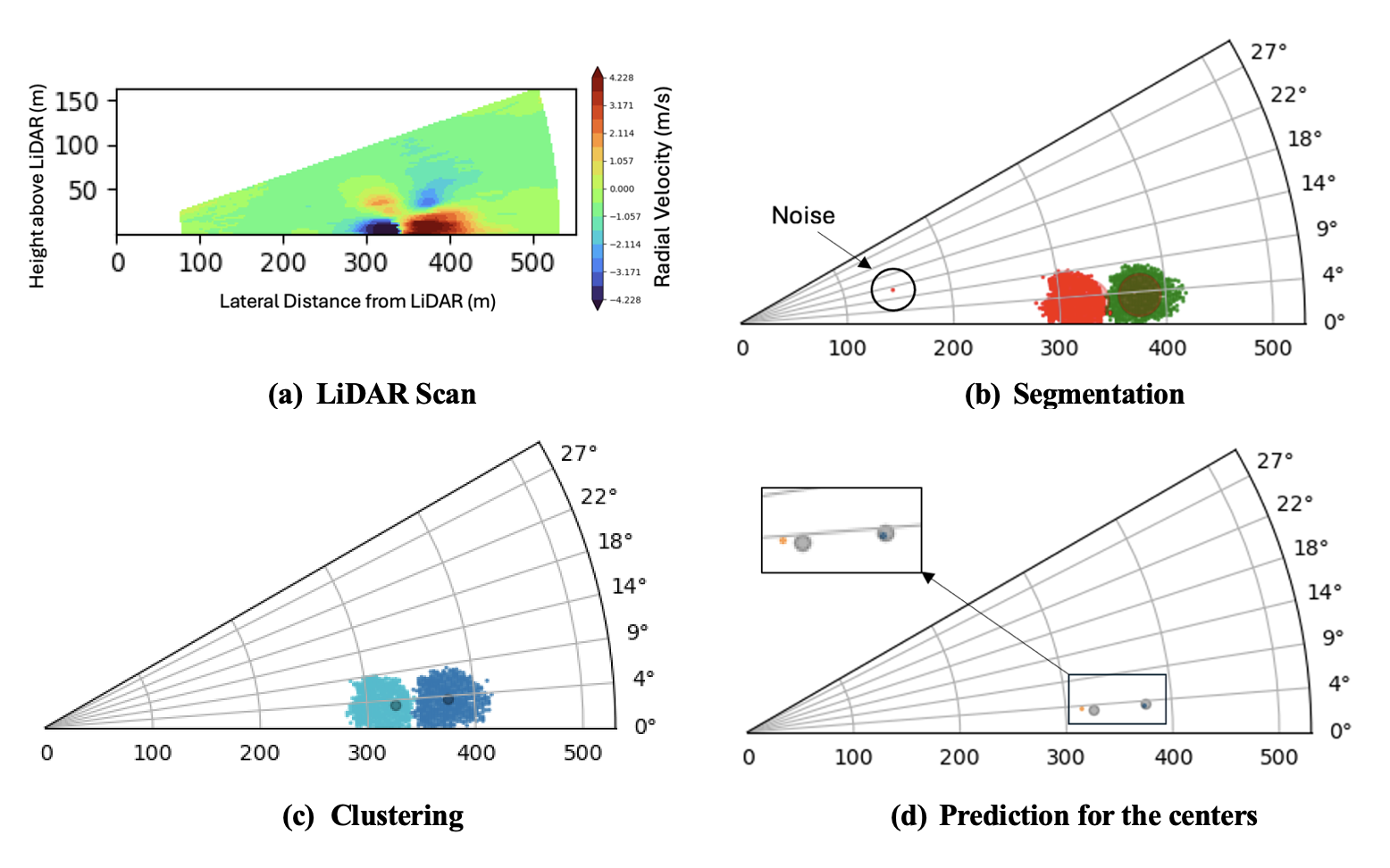}
  \caption{Overview of Our Approach.}
  \label{fig:overview}
  \Description[The four stages of our approach]{The four stages of our approach: data preprocessing, segmentation, clustering, and prediction of the centers}
\end{figure*}

\subsection{Explainability for 3D Point Clouds}

Explainability in point cloud analysis remains a frontier with limited exploration. While approaches like PointNet and DGCNN have pushed the boundaries of accuracy and efficiency, 
their explainability has still been less investigated. 
Although recent studies have begun to investigate explainability through gradient-based methods \cite{Grad3D3_ZCYLR19,Grad3D2_GWY20,Grad3D_TSS22} and surrogate methods \cite{LIME3D_TK22}, other techniques such as perturbation-based explanations \cite{fong2017interpretable} remain unexplored for models analyzing 3D point clouds.

Furthermore, conventional explanation methods like gradient-based and surrogate approaches are not ideally suited for segmentation tasks. These methods typically produce a heatmap to emphasize key regions in a scan. However, the output from segmentation models inherently consists of highlighted scan regions, rendering such heatmaps redundant. %
Consequently, perturbation-based explanation methods, which have already been widely applied in the computer vision \cite{fong2017interpretable} and graph domains \cite{ying2019gnnexplainer}, emerge as particularly fitting for this study. They offer a more relevant way to interpret and understand the segmentation results by directly manipulating the data and observing the impact on the model’s output.

\section{Methods}
\subsection{Overview}

The objective of our work is to accurately determine the geographic centers of wake vortex cores. As highlighted in Section~\ref{subsection:ml_wv}, prior research on machine learning-based wake vortex detection typically transforms LiDAR scans into images before applying machine learning models such as GNN and YOLO for prediction purposes. Contrary to these approaches, our study directly utilizes models designed for 3D point clouds on the LiDAR scans.

Our methodology encompasses several key steps, as depicted in Fig.~\ref{fig:overview}. The process starts with preprocessing LiDAR scans, where, in Fig.~\ref{fig:overview} (a), colors denote radial velocity—indicative of a point's movement direction relative to the LiDAR scanner, with blue for approaching points and red for receding ones. This coloring reflects the wake vortices' opposite rotations, resulting in the port vortex (right side) appearing with a blue top and red bottom, and vice versa for the starboard vortex (left side). The preprocessing steps include calculating the position of a measurement point in Cartesian coordinates and expanding the labels to fit segmentation models. After preprocessing, 3D point cloud segmentation models are utilized to detect wake vortices (b). Clustering algorithms then remove outlying measurement points (the noise in b) from the main clusters (c). Following this, we calculate the geographic centers of these clusters for our predictions (d). We also incorporate a novel perturbation-based explanation method to clarify the segmentation model's predictions. These explanations, essential for neural network-based models, are not illustrated in Fig.~\ref{fig:overview} as they are performed post-prediction. In the following section, we outline the key steps in more detail.

\subsection{Segmentation}

Given that the LiDAR scans retain their original polar coordinate format, it is essential to avoid possible scaling issues when transforming to other coordinate systems. Therefore, it is crucial to utilize models adept at processing points within these coordinates while minimizing preprocessing efforts. 
To this end, three distinct models—PointNet \cite{PointNet_QSMG17}, PointNet++ \cite{PointNet++_QYSG17}, and DGCNN \cite{DGCNN_WSLSBS19}—were selected for their unique approaches to utilizing spatial features within point clouds. These models, designed to operate directly on the point clouds, allow us to compute the geographic center of the segmented wake vortex once a segmentation is made.

The first model, PointNet \cite{PointNet_QSMG17}, treats each point individually, applying the same processing to every point to extract features. This model is revolutionary for its ability to learn global features through a symmetric function, ensuring invariance to point order. PointNet's architecture allows it to directly consume raw point clouds and output classifications or segmentations without requiring intricate preprocessing or the conversion of data into Cartesian coordinates.

PointNet++ \cite{PointNet_QSMG17}, an extension of PointNet, introduces a hierarchical approach to processing point sets. It builds upon the foundation laid by PointNet by incorporating local features at multiple scales. PointNet++ segments the point cloud into overlapping local regions, processes each region independently to capture local structures at various scales, and hierarchically aggregates these features. This method enhances the model's sensitivity to fine-grained patterns and structures within the data, making it particularly adept at dealing with the complexity and variability inherent in point clouds.

Dynamic Graph CNN (DGCNN) \cite{DGCNN_WSLSBS19} further innovates by dynamically generating graphs based on point proximity. Unlike traditional convolutional networks that operate on fixed grid structures, DGCNN constructs a graph for each point cloud layer, with edges connecting points based on their spatial relationships. This approach allows for the adaptive learning of features, taking into account the varying densities and arrangements of points within the cloud. For our purposes, the DGCNN model underwent slight modifications and reductions to better align with the unique characteristics of the LiDAR data in polar coordinates.

\subsection{Clustering}

To determine the position of the vortices, the segmentations are then refined and organized using various clustering algorithms, which play a crucial role in enhancing the accuracy of the results by removing outliers. Three primary clustering methods were employed: Agglomerative Clustering \cite{ward1963hierarchical}, DBSCAN \cite{ester1996density}, OPTICS \cite{ankerst1999optics}, alongside a simplified approach derived from DBSCAN. Each method offers a unique mechanism for grouping data points based on their spatial relationships and density, facilitating the identification of wake vortices from the segmented LiDAR data.

Agglomerative Clustering \cite{ward1963hierarchical} operates on a bottom-up approach, where each data point starts as its own cluster, and pairs of clusters are merged as one moves up the hierarchy. DBSCAN (Density-Based Spatial Clustering of Applications with Noise) \cite{ester1996density} groups together closely packed points, marking as outliers those that lie alone in low-density regions. This algorithm is distinguished by its ability to form clusters of arbitrary shape and size, relying on two key parameters: the minimum number of points required to form a cluster and the maximum distance between two points for them to be considered part of the same cluster. These assumptions reflect the nature of the vortex structure well, since it has to cover a certain amount of points to qualify as a vortex and it is confined to a certain area in space. OPTICS (Ordering Points To Identify the Clustering Structure) \cite{ankerst1999optics} extends the DBSCAN concept by addressing its sensitivity to the input parameters. Instead of relying on a single threshold for cluster formation, OPTICS uses an ordering of points based on their spatial density relationships, allowing for the identification of clusters across varying density scales. 

After these clustering algorithms are applied, the center point of each cluster can be determined, effectively pinpointing the location of each wake vortex. This methodology is capable of managing an unspecified number of wake vortices, without the need for prior knowledge about the aircraft's presence in the airspace. 

\subsection{Explanation}

To ensure that segmentation models are understandable and trustworthy for regulators, air traffic controllers, and pilots -- especially in non-standard situations -- it is crucial that the decisions of the AI models can be easily interpreted. 
Unlike the previous image-based approaches \cite{wartha2022characterizing, stephan2023artificial}, 
segmentation helps to identify specific critical points within the glide path and flags potential wake vortices in an understandable way. 
To validate the method's reliability, we create a novel perturbation-based method and apply it to the LiDAR data. 
This involves altering the data around a wake vortex by (1) masking out, (2) moving, or (3) swapping points that belong to the core regions of the wake vortices and then reassessing the modified scan. 
The effectiveness of the method is demonstrated when the altered regions are not identified as wake vortices. Meanwhile, the areas surrounding them are identified correctly, as the remaining points still suggest the presence of a wake vortex.
This approach also holds true when a wake vortex position is modified, confirming the method's accuracy in detecting wake vortices based on the spatial distribution of points.

In our masking approach, we obscure the core region by adjusting the radial velocity of points within it to the scan's average radial velocity. This process aims to eliminate segmentations in the masked areas, with new segmentations expected to form around these regions instead. For the moving strategy, we shift the core regions of the wake vortices to different locations, substituting the radial velocity of the original regions with the scan's average. 
This moving strategy should lead to the absence of segmentations in the original core regions, with the correct type (port or starboard) of segmentations emerging at the relocated sites. 
In the swapping method, we interchange the positions of the port and starboard vortices. Given the distinct characteristics of each vortex, this maneuver is anticipated to result in a segmentation where the features of the two vortices intermingle.

The goal of the perturbation-based explanation method is to prove the model's reliability, thereby ensuring that domain experts, who depend on these models for vital decision-making, can have confidence in the model's dependability. When predictions adjust in response to perturbations, it signifies that the key features identified by the machine learning model align with what domain experts also deem critical, reinforcing trust in the model's efficacy.

\begin{table*}[tb] %
    \caption{Comparison of Recall (\%), Mean Error (m) and Mean Error without False Negatives (m) of different model architectures}
    \centering
    \begin{tabular}{l l c c c c c c}
    \toprule
    \multirow{2}{*}{} & \multirow{2}{*}{Architecture} & \multicolumn{3}{c}{Measurement Data} & \multicolumn{3}{c}{Proxy Data}\\
    \cmidrule(lr){3-5}\cmidrule(lr){6-8}
    & & Recall (\%) & ME (m) & ME w/o fn (m) & Recall (\%) & ME (m) & ME w/o fn (m) \\
    \midrule
    \multirow{2}{*}{Image Processing} & CNN & 95.28 & 33.08 & 21.26 & 96.78 & 19.46 & 12.03 \\
    & YOLO & 94.00 & 22.44 & 14.59 & 95.43 & 15.22 & 10.57 \\
    \midrule
    \multirow{3}{*}{Segmentation} & PointNet & 95.48 & 24.54 & 10.88 & 98.65 & 17.53 & 15.86\\
    & PointNet++ & 91.33 & 33.33 & 11.19 & 96.51 & 21.94 & 11.32 \\
    & \textbf{DGCNN} & \textbf{99.50} & 10.09 & 8.73 & \textbf{98.87} & 10.34 & 5.25\\
    \midrule
    \multirow{3}{*}{Segmentation \& Clustering} & \textbf{DGCNN \& Agglo.} & \textbf{99.50} & \textbf{9.60} & \textbf{6.33} & \textbf{98.87} & \textbf{7.51} & \textbf{3.51}\\
    & DGCNN \& DBSCAN & 98.87 & 11.42 & 8.71 & 97.86 & 13.58 & 5.04 \\
    & DGCNN \& OPTICS     & 98.74 & 15.80 & 12.57 & 97.75 & 20.70 & 12.09\\
    \bottomrule
    \end{tabular}
    \label{tab:Architectures}
\end{table*}

\section{Experiments}

\subsection{Datasets and Experimental Setup}

\paragraph{Datasets} We use two datasets in our study. The primary dataset \cite{holzapfel2021mitigating}, obtained from Vienna International Airport, includes LiDAR scans that capture wake vortices with radial velocity measurements, featuring details like aircraft type and timestamps. The secondary dataset comprises synthetic data \cite{wartha2021characterizing}, devoid of background noise and is missing specific details such as aircraft type and timestamps. Both datasets can reveal up to three wake vortices per scan, as vortices from previous airplanes may persist in the LiDAR's scanning range for extended periods. Each vortex is classified as either port or starboard based on its rotational direction. 
Despite their differences, the datasets share key features and prediction objectives. 
In both datasets, ground truth is established solely by labeling the centers of the wake vortices. Labels originate from a physical wake vortex characterization algorithm named the Radial Velocity (RV) method \cite{smalikho2015method}.  
It should be kept in mind that labels from the RV method are by no means perfect and carry inherent errors. Investigations quantifying these errors, particularly under turbulent atmospheric conditions, have recently been conducted \cite{wartha2023investigating}. 

\paragraph{3D Point Cloud Construction} The operational parameters of the LiDAR instrument are crucial for data collection. Three key measurements are obtained from the LiDAR scans, which are essential for data preparation: (1) Elevation Angle $\varphi$, (2) Range Gate $R$, and (3) Radial Velocity $V_r$ (see Fig.~\ref{fig:wake-vortices-slices}). The elevation angle measures the beam's deviation from the ground, while the range gate denotes the distance from the LiDAR scanner to each measurement point. Utilizing elevation and range gate values allows for the determination of each measurement point's position in polar coordinates. Additionally, radial velocity quantifies the average velocity within each range gate—note that LiDAR cannot deliver point measurements—which, along with intensity, enables the construction of a 3D point cloud from the coordinates and intensities of these points. The prediction targets include identifying the central position of each wake vortex and distinguishing its type (Port or Starboard).

\paragraph{Labels} Addressing the challenge of accurately determining wake vortex positions, we adapt segmentation models for 3D point cloud analysis. These models require precise preprocessing of the datasets to suit the task. Because the datasets only label the vortex centers, yet a segmentation model requires labeling of the entire vortices, adjustments for the labels are necessary. Given the complexity of the wake vortices, we have opted for a simplified approach to label the data. To facilitate accurate analysis, three classes have been established: a background class, a port wake vortex class, and a starboard wake vortex class. For ground truth determination, points within a fixed radius $r$ surrounding a wake vortex's center are allocated to the class corresponding to that vortex's type. Data points not part of a wake vortex are labeled as background. The optimal value for the radius $r$ was determined through a series of trials, with the selection based on the configuration that yielded the most favorable outcomes, which was then utilized for subsequent experiments.

Points within a 25m radius of the actual center of a wake vortex were classified as part of the wake vortex. To ensure uniformity across scans with varying point counts, 12,000 points were randomly selected from each scan. The radial velocity values of the proxy data were normalized to fall within the [0,1] range. 

\paragraph{Models and Experimental Setup} The models underwent training for different durations: 50 epochs for DGCNN and 100 epochs for PointNet and PointNet++. Cross-entropy loss was employed as the loss function. The ADAM optimizer, with a learning rate of 0.001, was used for optimization. Due to the complexity of calculating dynamic graphs, DGCNN had a batch size of 4, while PointNet and PointNet++ used a batch size of 16. For DGCNN, the k-nearest neighbors parameter was set to 20 for constructing dynamic graphs. Both datasets were balanced with equal sample sizes. The training set comprised 2,000 LiDAR scans, while the test set included 500 unseen LiDAR scans. Each test set contained a total of 898 wake vortices. To further validate our approach, 
we use CNN \cite{lecun1998gradient} and YOLO \cite{redmon2016you}, current state-of-the-art methods, 
as additional baselines. These methods require us to first transform the 3D point clouds into 2D images by visualizing the 3D point clouds in Cartesian coordinates, consistent with previous studies \cite{wartha2022characterizing, stephan2023artificial}.

Recall was assessed for each scan to determine if the wake vortices were correctly identified. The mean error was calculated based on the distance between the predicted vortex center and the actual center. In instances where a wake vortex was missed (False Negative), the error was measured from the origin point (0,0) in the Cartesian coordinate system.

\subsection{Segmentation and Clustering Results}

\begin{figure}[tb]
    \centering
    \begin{subfigure}{0.33\textwidth}
        \includegraphics[width=\textwidth, trim={70 78 100 90},clip]{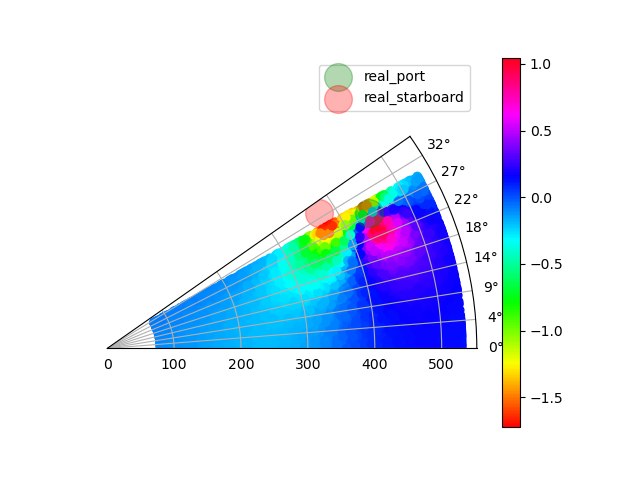}
        \caption{Wake Vortices on the borders}
    \end{subfigure}
    \hfill
    \begin{subfigure}{0.33\textwidth}
        \includegraphics[width=\textwidth, trim={70 78 100 90},clip]{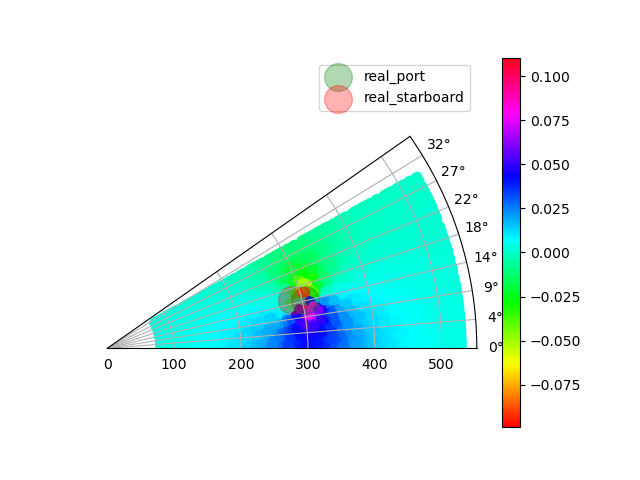}
        \caption{Mislabeled Wake Vortices}
    \end{subfigure}
    \caption{Example scans for poor predictions. The circles correspond to the labeled wake vortices.}
    \label{fig:problems}
    \Description[2 Examples of poor predictions]{In the top figure, the 2 vortices are in the corner of the scan; In the bottom, the only vortex is mislabeled as 2 vortices}
\end{figure}

Table~\ref{tab:Architectures} presents the evaluation results for DGCNN and the baseline models—CNN, YOLO, PointNet, and PointNet++—on the task of predicting the centers of wake vortices (semantic segmentation for 3D point clouds). The performance was assessed using recall, mean error, and adjusted mean error. DGCNN demonstrated superior performance across all metrics in both datasets. To enhance the accuracy of the mean error calculation, we excluded false negatives, thus refining the evaluation metric. This adjustment addresses labeling inaccuracies within the datasets, particularly for vortices near scan boundaries or those mislabeled (see Fig.~\ref{fig:problems}). A manual investigation of the samples revealed that, for the DGCNN model, all false negatives were associated with labeling issues. This indicates that the model is robust against such errors, highlighting the need for further exploration in future studies.

Given DGCNN's superior performance, it was further combined with three clustering algorithms to refine the predictions. Results, shown in Table~\ref{tab:Architectures}, indicate all clustering methods outperformed the base results of PointNet and PointNet++ alone. However, only Agglomerative clustering enhanced DGCNN's performance, while DBSCAN and OPTICS were less effective, especially in separating closely located vortices, thus affecting recall. Agglomerative Clustering not only preserved recall but also refined center predictions by filtering out distant noise points, and effectively ignoring small, distant clusters, as illustrated in Fig.~\ref{fig:overview} (b \& c). Our evaluation results demonstrate DGCNN's accurate segmentation and the effectiveness of Agglomerative Clustering in segmentation refinement.

\begin{figure*}[tb] %
    \centering
    \includegraphics[width=0.9\textwidth]{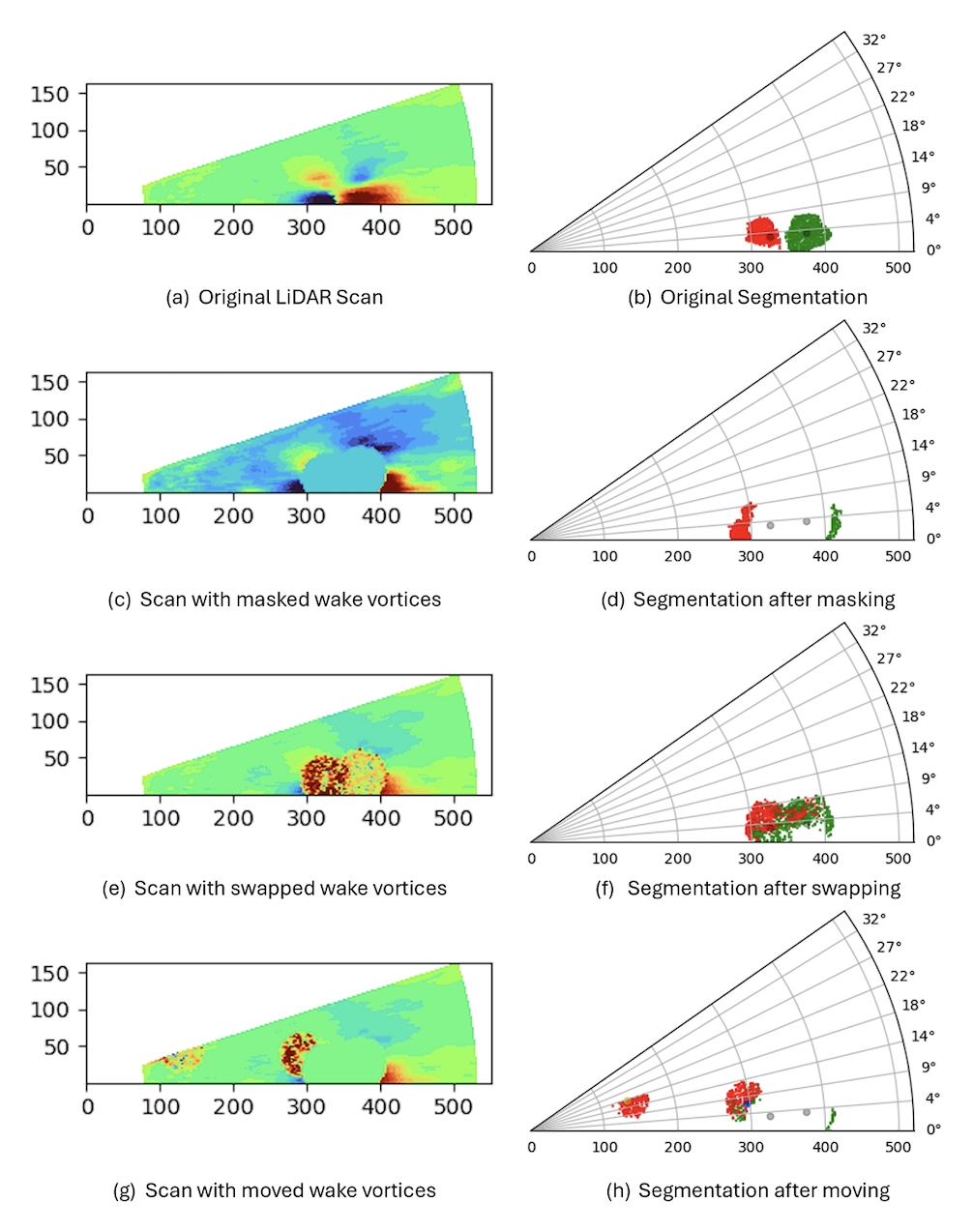}
    \caption{Examples for Perturbation Methods}
    \label{fig:perturbations}
    \Description[Different methods for measuring the explainability of the model]{From top to bottom: original scan and segmentation, scan and segmentation after masking, scan and segmentation after swapping, scan and segmentation after moving}
\end{figure*}

\subsection{Explainability Study}

We apply three distinct perturbation techniques in our evaluation: (1) Masking, (2) Swapping and (3) Moving the core regions of the wake vortices. Figure~\ref{fig:perturbations} illustrates the changes in the predictions after these perturbations compared to the original LiDAR scans and their subsequent segmentations. As exemplified in Figure~\ref{fig:perturbations}, this approach can provide an explanation for the developed method. When the center of a wake is masked out (c \& d), only the area around it is segmented as a wake vortex. This proves empirically that the model recognizes the difference between noise and wake vortices, especially since the area around the hidden center is still segmented correctly. The swapped wake centers (e \& f) show that the shifted points are sufficient to identify the associated wake class, but are largely obscured by the surrounding points. As a result, fragments of the other wake vortex class are created during segmentation when the wake vortex centers are swapped, while the surrounding points are still assigned to the original class. A similar picture emerges when moving the wake vortex centers (g \& h). Here too, the points surrounding the original center remain segmented as wake vortices, while a wake vortex can usually also be recognized at the moved target point.

\section{Conclusions}

In this paper, we investigated advanced 3D point cloud segmentation and clustering techniques to accurately identify the core centers of wake vortices, a crucial factor for air traffic safety. Using two datasets, one with real LiDAR scans from Vienna Airport and another with synthetic data, we demonstrated the superiority of the Dynamic Graph Convolutional Neural Network (DGCNN) model over other architectures such as PointNet and PointNet++. Our evaluation results showed DGCNN's improved performance in wake vortex detection, especially when combined with agglomerative clustering algorithms. This combination not only preserved recognition rates but also minimized prediction errors, refining the detection process. 
Our results also showed that models designed specifically for 3D point cloud segmentation significantly outperform 2D image processing models, such as CNN and YOLO.

Our experiments highlighted the challenges of wake vortex detection, such as the effects of vortices at the edge of the scan or inaccurately labeled data. By applying perturbation-based explanation techniques, we gained insights into the model's operational dynamics and confirmed its ability to distinguish between noise and true wake vortices. The exploration of perturbations—masking, swapping, and displacement of vortex core regions—provided a deeper understanding of our methodology's effectiveness in detecting and classifying wake vortices.

Our work pushed the boundaries of wake vortex detection using state-of-the-art 3D point cloud processing techniques and set a new benchmark for the interpretability and explainability of such models. The successful application of these models to real and synthetic datasets demonstrated the robustness and adaptability of our approach, enhancing both safety and operational efficiency. 

\section*{Acknowledgment}
This work was partially supported by the ``mFUND'' funding program of the German Federal Ministry for Digital and Transport (BMDV) as part of the project ``KIWI'' (grant number 19F1137B). 
Additionally, it was partially supported by the German Federal Ministry of Education and Research (BMBF) as part of the Software Campus project ``LLM4Edu'' (grant number 01IS23070).
We thank Grigory Rotshteyn for his insightful contributions  to this work. 

\bibliographystyle{ACM-Reference-Format}
\bibliography{references}

\end{document}